\documentclass[a4paper,twoside]{article}

\usepackage{epsfig}
\usepackage{subfigure}
\usepackage{calc}
\usepackage{amssymb}
\usepackage{amstext}
\usepackage{amsmath}
\usepackage{amsthm}
\usepackage{multicol}
\usepackage{hyperref}
\usepackage{pslatex}
\usepackage{apalike}
\usepackage{soul}
\usepackage{SCITEPRESS}     
\usepackage{color}
\begin{document}

\title{Exploring applications of deep reinforcement learning  
\\ for real-world autonomous driving systems}

\author{\authorname{ Victor Talpaert\sup{1, 2}, Ibrahim Sobh\sup{3}, B Ravi Kiran\sup{2}, Patrick Mannion\sup{4}, \\ Senthil Yogamani\sup{5}, Ahmad El-Sallab\sup{3} and Patrick Perez\sup{6}}
\affiliation{\sup{1, 2} U2IS, ENSTA ParisTech, Palaiseau, France, 
\sup{2} AKKA Technologies, Guyancourt, France, 
\sup{3} Valeo Egypt, Cairo, 
\sup{5} Valeo Vision Systems, Ireland, 
\sup{4} Galway-Mayo Institute of Technology, Ireland, 
\sup{6} Valeo.ai, France}
}



\abstract{Deep Reinforcement Learning (DRL) has become increasingly powerful in recent years, with notable achievements such as Deepmind's AlphaGo. It has been successfully deployed in commercial vehicles like Mobileye's path planning system. However, a vast majority of work on DRL is focused on toy examples in controlled synthetic car simulator environments such as TORCS and CARLA. In general, DRL is still at its infancy in terms of usability in real-world applications. Our goal in this paper is to encourage real-world deployment of DRL in various autonomous driving (AD) applications. We first provide an overview of the tasks in autonomous driving systems, reinforcement learning algorithms and applications of DRL to AD systems. We then discuss the challenges which must be addressed to enable further progress towards real-world deployment. 
}

\onecolumn \maketitle \normalsize \vfill

\section{Introduction} 
\label{intro}

Autonomous driving (AD) is a challenging application domain for machine learning (ML). Since the task of driving ``well'' is already open to subjective definitions, it is not easy to specify the correct behavioural outputs for an autonomous driving system, nor is it simple to choose the right input features to learn with.
Correct driving behaviour is only loosely defined, as different responses to similar situations may be equally acceptable. ML-based control policies have to overcome the lack of a dense metric evaluating the driving quality over time, and the lack of a strong signal for expert imitation. Supervised learning methods do not learn the dynamics of the environment nor that of the agent \cite{sutton2018reinforcement}, while reinforcement learning (RL) is formulated to handle sequential decision processes, making it a natural approach for learning AD control policies. 

In this article, we aim to outline the underlying principles of DRL for applications in AD. Passive perception which feeds into a control system does not scale to handle complex situations. DRL setting would enable active perception optimized for the specific control task. The rest of the paper is structured as follows. Section \ref{background} provides background on the various modules of AD, an overview of reinforcement learning and a summary of applications of DRL to AD. Section \ref{challenges} discusses challenges and open problems in applying DRL to AD. Finally, Section \ref{conclusion} summarizes the paper and provides key future directions.

\section{Background}
\label{background}

\begin{figure*}
\centering
\includegraphics[width=0.95\linewidth]{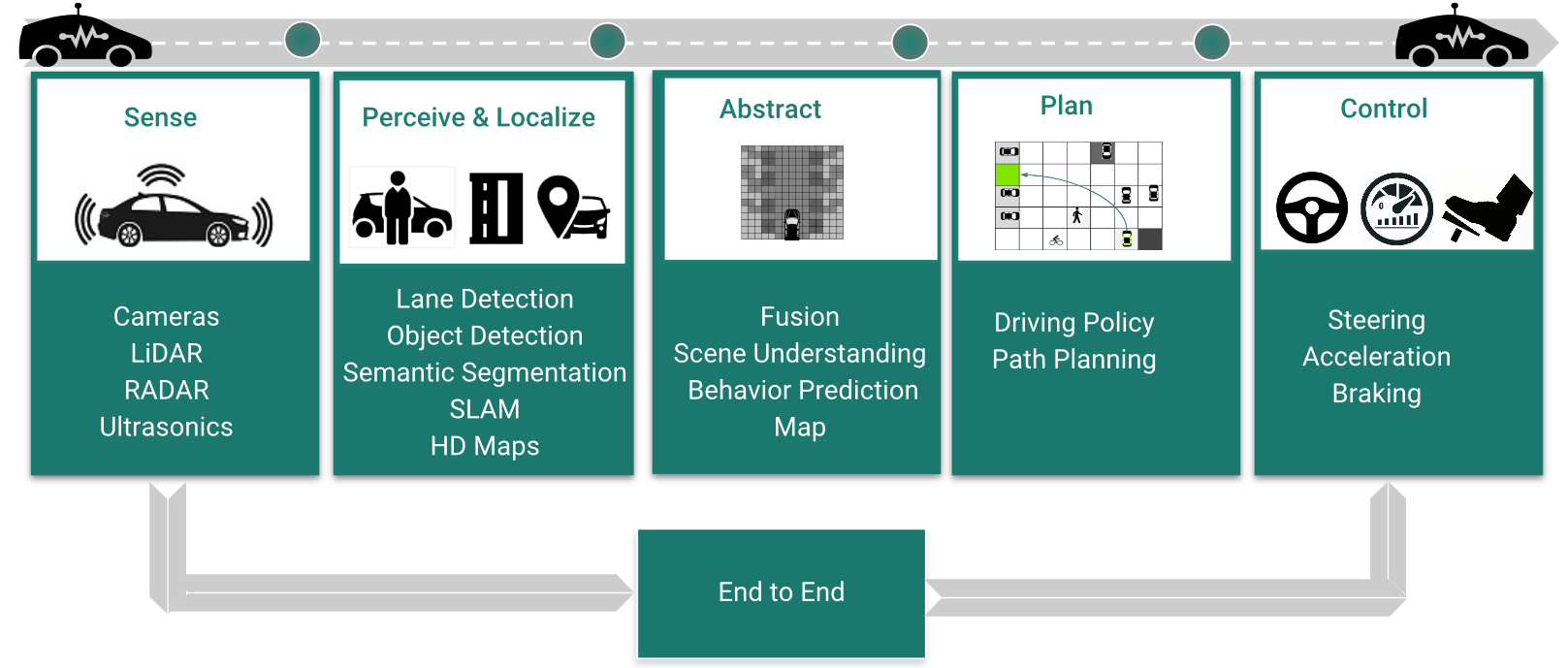}
\caption{Fixed modules in a modern autonomous driving systems : \textbf{Sensor infrastructure} notably include Cameras, Radars, LiDARs (the Laser equivalent of Radars) or GPS-IMUs (GPS and Inertial Measurement Units provide an instantaneous position); their raw data is considered low level. This dynamic data is often processed into higher level descriptions as part of the Perception module. \textbf{Perception} estimates positions of: location in lane, cars, pedestrians, traffic lights and other semantic objects among other descriptors. It also provides road occupation over a larger spatial \& temporal scale by combining maps.
\textbf{Mapping} Localised high definition maps (HD maps) provide centrimetric reconstruction of buildings, static obstacles and dynamic objects, which frequently are crowdsourced. \textbf{Scene understanding} provide the high-level scene comprehension that includes detection, classification \& localisation tasks, feeding the driving policy/planning module. \textbf{Path Planning} predicts the future actor trajectories and manoeuvres. A static shortest path from point $A$ to point $B$ with dynamic traffic information constraints are employed to calculate the path. \textbf{Vehicle control} orchestrates the high level orders for motion planning using simple closed loop systems based on sensors from the Perception task.}
\label{ADflow}
\end{figure*}

The software architecture for Autonomous Driving systems comprises of the following high level tasks: Sensing, Perception, Planning and Control. While some \cite{DBLP:journals/corr/BojarskiTDFFGJM16} achieve all tasks 
in one unique module, others do follow the logic of one task - one module \cite{DBLP:journals/corr/PadenCYYF16}. Behind this separation rests the idea of information transmission from the sensors to the final  stage of actuators and motors as illustrated in Figure \ref{ADflow}. While an End-to-End approach would enable one generic encapsulated solution, the modular approach provides granularity for this multi-discipline problem, as separating the tasks in modules is the \textit{divide and conquer} engineering approach.

\subsection{Review of Reinforcement learning}

Reinforcement Learning (RL) \cite{sutton2018reinforcement} is a family of algorithms which allow agents to learn how to act in different situations. In other words, how to establish a map, or a policy, from situations (states) to actions which maximize a numerical reward signal. RL
has been successfully applied to many different fields such as helicopter
control \cite{naik1992meta}, traffic signal control \cite{Mannion2016Experimental},
electricity generator scheduling \cite{Mannion2016Multi}, water resource management
\cite{Mason2016Applying}, playing relatively simple Atari games
\cite{mnih2015human} and mastering a much more complex game of Go
\cite{silver2016mastering}, simulated continuous control problems
\cite{lillicrap2015continuous}, \cite{schulman2015trust}, and controlling robots
in real environments \cite{levine2016end}. 

RL algorithms may learn estimates of state values, environment models or policies. In real-world applications, simple tabular representations of estimates are not scalable. Each additional feature tracked in the state leads to an exponential growth in the number of estimates that must be stored \cite{sutton2018reinforcement}. 

Deep Neural Networks (DNN) have recently been applied as function approximators for RL agents, allowing agents to generalise knowledge to new unseen situations, along with new algorithms for problems with continuous state and action spaces. Deep RL agents can be value-based: DQN \cite{mnih2013playing}, Double DQN \cite{van2016deep}, 
policy-based: TRPO
\cite{schulman2015trust}, PPO \cite{schulman2017proximal}; or actor-critic: DPG
\cite{silver2014deterministic}, DDPG \cite{lillicrap2015continuous}, A3C
\cite{mnih2016asynchronous}. Model-based RL agents attempt to build an environment model, e.g. Dyna-Q \cite{sutton1990integrated}, Dyna-2 \cite{silver2008sample}. 
Inverse reinforcement learning (IRL) aims to estimate the reward function given examples of agents actions, sensory input to the agent (state), and the model of the environment \cite{abbeel2004apprenticeship}, \cite{ziebart2008maximum}, \cite{finn2016guided},  \cite{ho2016generative}.


\subsection{DRL for Autonomous Driving}

\begin{table*}[htbp]
\centering
 \begin{tabular}{||p{7.5cm}|p{7cm}||} 
 \hline
 \textbf{Dataset} & \textbf{Description} \\ [0.5ex] 
 \hline\hline
 Berkeley Driving Dataset \cite{BDD_XuGYD17} & Learn driving policy from demonstrations \\ 
 Baidu's ApolloScape & Multiple sensors \& Driver Behaviour Profiles \\
 Honda Driving Dataset \cite{ramanishka2018toward} &  4 level annotation (stimulus, action, cause, and attention objects) for driver behavior profile.\\
 \hline\hline
 \textbf{Simulator} & \textbf{Description} \\ [0.5ex] 
 \hline
 CARLA \cite{Dosovitskiy17} & Urban Driving Simulator with Camera, LiDAR, Depth \& Semantic segmentation\\
 Racing Simulator TORCS \cite{wymann2000torcs} & Testing control policies for vehicles\\
 AIRSIM \cite{shah2018airsim} & Resembling CARLA with support for Drones\\
 GAZEBO \cite{Koenig04designand_GAZEBO} & Multi-robo simulator for planning \& control\\ 
\hline
\end{tabular}
\caption{A collection of datasets and simulators to evaluate AD algorithms.}
\label{table_Dataset}
\end{table*}

\noindent
In this section, we visit the different modules of the autonomous driving system as
shown in Figure \ref{ADflow} and describe how they are achieved using classical RL 
and Deep RL methods. A list of datasets and simulators for AD tasks is presented in Table \ref{table_Dataset}. \\

\noindent \textbf{Vehicle control}: Vehicle control classically has been achieved with predictive control approaches such as Model Predictive Control (MPC) \cite{DBLP:journals/corr/PadenCYYF16}.
A recent review on motion planning and control task can be found by authors \cite{schwarting2018planning}.
Classical RL methods are used to perform optimal control in stochastic settings the Linear Quadratic Regulator (LQR) in linear regimes and iterative LQR (iLQR) for non-linear regimes are utilized. More recently, random search over the parameters for a policy network can perform as well as LQR \cite{mania2018simple}.

One can note recent work on DQN which is used in \cite{yu2016deep} for simulated autonomous vehicle control where different reward functions are examined to produce specific driving behavior. The agent successfully learned the turning actions and navigation without crashing. In \cite{sallab2016end} DRL system for lane keeping assist is introduced for discrete actions (DQN) and continuous actions (DDAC), where the TORCS car simulator is used and concluded that, as expected, the continuous actions provide smoother trajectories, and the more restricted termination conditions, the slower convergence time to learn. Wayve, a recent startup, has recently demonstrated an application of DRL (DDPG) for AD using a full-sized autonomous vehicle \cite{kendall2018learning}. The system was first trained in simulation, before being trained in real time using onboard computers, and was able to learn to follow a lane, successfully completing a real-world trial on a 250 metre section of road. \\

\noindent \textbf{DQN for ramp merging}: The AD problem of ramp merging is tackled in \cite{wang2017formulation}, where DRL is applied to find an optimal driving policy using LSTM for producing an internal state containing historical driving information and DQN for Q-function approximation.
\textbf{Q-function for lane change}: A Reinforcement Learning approach is proposed in \cite{wang2018reinforcement} to train the vehicle agent to learn an automated lane change in a smooth and efficient behavior, where the coefficients of the Q-function are learned from neural networks. \\

\noindent \textbf{IRL for driving styles}: Individual perception of comfort from demonstration is proposed in \cite{kuderer2015learning}, where individual driving styles are modeled in terms of a cost function and use feature based inverse reinforcement learning to compute trajectories in vehicle autonomous mode.
Using Deep Q-Networks as the refinement step in IRL is proposed in \cite{sharifzadeh2016learning} to extract the rewards. While evaluated in a simulated autonomous driving environment, it is shown that the agent performs a human-like lane change behavior. \\

\noindent \textbf{Multiple-goal RL for overtaking}: In \cite{ngai2011multiple} a multiple-goal reinforcement learning (MGRL) framework is used to solve the vehicle overtaking problem. This work is found to be able to take correct actions for overtaking while avoiding collisions and keeping almost steady speed. \\

\noindent \textbf{Hierarchical Reinforcement Learning(HRL)}: 
Contrary conventional or flat RL, HRL refers to the decomposition of complex agent behavior using temporal abstraction, such as the options framework \cite{barto2003recent}.
The problem of sequential decision making for autonomous driving with distinct behaviors is tackled in \cite{chen2018deep}. A hierarchical neural network policy is proposed where the network is trained with the Semi-Markov De\noindent cision Process (SMDP) though the proposed hierarchical policy gradient method. The method is applied to a traffic light passing scenario, and it is shown that the method is able to select correct decisions providing better performance compared to a non-hierarchical reinforcement learning approach. An RL-based hierarchical framework for autonomous multi-lane cruising is proposed in \cite{nosrati2018towards} and it is shown that the hierarchical design enables significantly better learning performance than a flat design for both DQN and PPO methods. \\

\noindent \textbf{Frameworks} A framework for an end-end Deep Reinforcement Learning pipeline for autonomous driving is proposed in \cite{sallab2017deep}, where the inputs are the states of the environment and their aggregations over time, and the output is the driving actions. The framework integrates RNNs and attention glimpse network, and tested for lane keep assist algorithm.
In this section we reviewed in brief the applications of DRL and classical RL methods to different AD tasks.

\subsection{Predictive perception}
In this section, we review some examples of applications of IOC and IRL to predictive perception tasks. Given a certain instance where the autonomous agent is driving in the scene, the goal of predictive perception algorithms are to predict the trajectories or intention of movement of other actors in the environment. The authors \cite{djuric2018motion}, trained a deep convolutional neural 
network (CNN) to predict short-term vehicle trajectories, while accounting
for inherent uncertainty of vehicle motion in road traffic. Deep Stochastic IOC (inverse optimal control) RNN Encoder-Decoder (DESIRE) \cite{lee2017desire} is a framework used to estimate a distribution and not just a simple prediction of an agent's future positions. This is based on the context (intersection, relative position of other agents). \\

\noindent
\textbf{Pedestrian Intention:}  Pedestrian intents to cross the road, board another vehicle, or is the driver in the parked car going to open the door \cite{MLAV2017ICML} . Authors in \cite{bziebart2009planning} perform maximum entropy inverse optimal control to learn a generic cost function for a robot to avoid pedestrians. Authors \cite{pedestrian_prediction2012} used inverse optimal control to predict pedestrian paths by considering scene semantics.  \\

\noindent
\textbf{Traffic Negotiation:}  When in traffic scenarios involving multiple agents, policies learned require agents to negotiate movement in densely populated areas and with continuous movement. MobilEye demonstrated the use of the options framework \cite{shalev2016safe}.

\section{Practical Challenges} 
\label{challenges}
Deep reinforcement learning is a rapidly growing field. We summarize the frequent challenges such as sample complexity and reward formulation, that could be encountered in the design of such methods.

\subsection{Bootstrapping RL with imitation} 
Ability learning by imitation is used by humans to teach other humans new skills. Demonstrations usually focus on state space essential areas from the expert's point of view. Learning from Demonstrations (LfD) is significant especially in domains where rewards are sparse. In imitation learning, an agent learns to perform a task from demonstrations without any feedback rewards. The agent learns a policy as a supervised learning process over state-action pairs. However, high quality demonstrations are hard to collect, leading to sub-optimal policies \cite{atkeson1997robot}. Accordingly, LfD can be used to initialize the learning agent with a policy inspired by performance of an expert. Then, RL can be conducted to discover a better policy by interacting with the environment. Learning a model of the environment condensing LfD and RL is presented in \cite{abbeel2005exploration}. Measuring the divergence between the current policy and the expert for policy optimization is proposed in \cite{kang2018policy}. DQfD \cite{hester2017deep} pre-trains the agent and uses demonstrations by adding them into the replay buffer of the DQN and giving them additional priority. More recently, a training framework that combines LfD and RL for fast learning asynchronous agents is proposed in \cite{sobh2018fastlearning}.  


\subsection{Exploration Issues with Imitation} 
In some cases, demonstrations from experts are not available or even not covering the state space leading to learning a poor policy. 
One solution consists in using the Data Aggregation (DAgger) \cite{ross2010efficient} methods where the end-to-end learned
policy is run and extracted observation-action pair is again labelled by the
expert, and aggregated to the original expert observation-action dataset. Thus
iteratively collecting training examples from both reference and trained policies
explores more states and solves this lack of exploration. Following work on
Search-based Structured Prediction (SEARN) \cite{ross2010efficient}, Stochastic
Mixing Iterative Learning (SMILE) trains a stochastic stationary policy over
several iterations and then makes use of a \textit{geometric} stochastic mixing of the
policies trained. In a standard imitation learning scenario, the demonstrator is
required to \textit{cover} sufficient states so as to avoid unseen states during test.
This constraint is costly and requires frequent human intervention. Hierarchical imitation learning methods reduce the sample complexity of standard imitation learning by performing data aggregation by organizing the action spaces in a hierarchy \cite{le2018hierarchical}.

\subsection{Intrinsic Reward functions}
In controlled simulated environments such as games, an explicit reward signal is given to the agent along with its sensor stream. In real-world robotics and autonomous driving deriving, designing a \textit{good} reward functions is essential so that the desired behaviour may be learned. The most common solution has been reward shaping \cite{ng1999policy} and consists in supplying additional rewards to the agent along with that provided by the underlying MDP. Rewards as already pointed earlier in the paper, could be estimated by inverse RL (IRL) \cite{AbbeelNg2004}, which depends on expert demonstrations.

In the absence of an explicit reward shaping and expert demonstrations, agents can use intrinsic rewards or intrinsic motivation \cite{chentanez2005intrinsically} to evaluate if their actions were good. Authors \cite{pathak2017curiosity} define curiosity as the error in an agent’s ability to predict the consequence of its own actions in a visual feature space learned by a self-supervised inverse dynamics model. 
In \cite{burda2018large} the agent learns a next state predictor model from its experience, and uses the error of the prediction as an intrinsic reward. This enables that agent to determine what could be a useful behavior even without extrinsic rewards.

\subsection{Bridging the simulator-reality gap}
Training deep networks requires collecting and annotating a lot of data which is
usually costly in terms of time and effort. Using simulation environments enables
the collection of large training datasets. However, the simulated data do not have the same
data distribution compared to the real data. Accordingly, models trained on
simulated environments often fail to generalise on real environments.
\textit{Domain adaptation} allows a machine learning model trained on samples from
a source domain to generalise to a target domain. \textbf{Feature-level domain
adaptation} focuses on learning domain-invariant features. In work of
\cite{ganin2016domain}, the decisions made by deep neural networks are based on
features that are both discriminative and invariant to the change of domains.
\textbf{Pixel level domain adaptation} focuses on stylizing images from the source
domain to make them \textit{similar} to images of the target domain, based
on image conditioned generative adversarial networks (GANs). In
\cite{bousmalis2017unsupervised}, the model learns a transformation in the pixel
space from one domain to the other, in an unsupervised way. GAN is used to adapt
simulated images to look like as if drawn from the real domain. Both feature-level
and pixel-level domain adaptation combined in \cite{bousmalis2017using}, where the
results indicate that including simulated data can improve the vision-based
grasping system, achieving comparable performance with 50 times fewer real-world
samples. Another relatively simpler method is introduced in \cite{peng2017sim}, by
\textit{dynamics randomizing} of the simulator during training, policies are
capable of generalising to different dynamics without any training on the real
system.
\textit{RL with Sim2Real}: A model trained in virtual environment is shown to be workable in real environment \cite{pan2017virtual}. Virtual images rendered by a simulator environment are first segmented to scene parsing representation and then translated to synthetic realistic images by the proposed image translation network. The proposed network segments the simulated image input, and then generates a synthetic realistic images. Accordingly, the driving policy trained by reinforcement learning can be easily adapted to real environment.

World Models proposed in \cite{ha2018world,ha2018recurrent} are trained quickly in an unsupervised way, via a Variational AutoEncoder (VAE), to learn a compressed spatial and temporal representation of the environment leading to learning a compact policy. Moreover, the agent can train inside its own dream and transfer the policy back into the actual environment.

\subsection{Data Efficient and fast adapting RL} 
Depending on the task being solved, RL require a lot of observations to cover the state space.
Efficieny is usually achieved with imitation learning, reward shaping and transfer learning.
Readers are directed towards the survey on transfer learning in RL here
\cite{taylor2009transfer}. The primary motivation in transfer learning in RL is to reuse
previously trained policies/models for a source task, so as to reduce the current target task's
training time. Authors in \cite{liaw2017composing} study the policy composition
problem where composing previously learned basis policies, e.g., driving in different conditions
or over different terrain types, the goal is to be able to reuse them for a novel task that is a
mixture of previously seen dynamics by learning a meta-policy, that maximises the reward. Their
results show that learning a new policy for a new task takes longer than a meta-policy learnt using basis policies.

Meta-learning, or learning to learn, is an important approach towards versatile agents that can adapt quickly to news tasks.
The idea of having one neural network interact with another one for meta-learning has been applied in \cite{duan2016rl} and \cite{wang2016learning}.  
More recently, the Model-Agnostic Meta-Learning (MAML) is proposed in \cite{finn2017model}, where the meta-learner seeks to find an initialization for the parameters of a neural network, that can be adapted quickly for a new task using only few examples. Continuous adaptation in dynamically changing and adversarial scenarios is presented in \cite{al2017continuous} via a simple gradient-based meta-learning algorithm. Additionally, Reptile \cite{nichol2018first} is mathematically similar to first-order MAML, making it consume less computation and memory than MAML. 

\subsection{Incorporating safety in DRL for AD}
Deploying an autonomous vehicle after training directly could be dangerous. 
We review different approaches to incorporate safety into DRL algorithms.\\

\noindent \textbf{SafeDAgger} \cite{SafeDAgger_AAAI2017} introduces a safety policy that learns to predict the error made by a primary policy trained initially with the supervised learning approach, without querying a reference policy. An additional safe policy takes both the partial observation of a state and a primary policy as inputs, and returns a binary label indicating whether the primary policy is likely to deviate from a reference policy without querying it.  \\

\noindent \textbf{Multi-agent RL for comfort driving and safety}: In \cite{shalev2016safe}, autonomous driving is addressed as a multi-agent setting where the host vehicle applies negotiations in different scenarios; where balancing is maintained between unexpected behavior of other drivers and not to be too defensive. The problem is decomposed into a policy for learned desires to enable comfort of driving, and trajectory planning with hard constraints for safety of driving. \\

\noindent \textbf{DDPG and safety based control}: The deep reinforcement learning (DDPG) and safety based control are combined in \cite{xiong2016combining}, including artificial potential field method that is widely used for robot path planning. Using TORCS environment, the DDPG is used first for learning a driving policy in a stable and familiar environment, then policy network and safety-based control are combined to avoid collisions.  It was found that combination of DRL and safety-based control performs well in most scenarios. \\

\noindent \textbf{Negative-Avoidance for Safety}: In order to enable DRL to escape local optima, speed up the training process and avoid danger conditions or accidents, Survival-Oriented Reinforcement Learning (SORL) model is proposed in \cite{ye2017survival}, where survival is favored over maximizing total reward through modeling the autonomous driving problem as a constrained MDP and introducing Negative-Avoidance Function to learn from previous failure. The SORL model found to be not sensitive to reward function and can use different DRL algorithms like DDPG.

\subsection{Other Challenges}

\paragraph{Multimodal Sensor Policies}  
Modern autonomous driving systems constitute of multiple modalities \cite{el2018end}, for example Camera RGB, Depth, Lidar and others sensors. Authors in \cite{multimodal_policies2017} propose 
end-to-end learning of policies that leverages sensor fusion to reduced performance
drops in noisy environment and even in the face of partial sensor failure by using 
Sensor Dropout to reduce sensitivity to any sensor subset.

\paragraph{Reproducibility} State-of-the-art deep RL methods are seldom
reproducible. Non-determinism in standard benchmark environments, combined with variance intrinsic to the methods, can make reported results tough to interpret, authors discuss common issues and challenges \cite{aaai2018}. Therefore, in the future it would be helpful to develop standardized benchmarks for evaluating autonomous vehicle control algorithms, similar to the benchmarks such as KITTI \cite{geiger2012we} which are already available for AD perception tasks.

\section{Conclusion}
\label{conclusion}

\noindent AD systems present a challenging environment for tasks such as perception, prediction and control. DRL is a promising candidate for the future development of AD systems, potentially allowing the required behaviours to be learned first in simulation and further refined on real datasets, instead of being explicitly programmed. In this article, we provide an overview of AD system components, DRL algorithms and applications of DRL for AD. We discuss the main challenges which must be addressed to enable practical and wide-spread use of DRL in AD applications. Although most of the work surveyed in this paper were conducted on simulated environments, it is encouraging that applications on real vehicles are beginning to appear, e.g. \cite{kendall2018learning}. The key challenges in constructing a complete real-world system would require resolving key challenges such as safety in RL, improving data efficiency and finally enabling transfer learning using simulated environments. To add, environment dynamics are better modeled by considering the predictive perception of other actors. We hope that this work inspires future research and development on DRL for AD, leading to increased real-world deployments in AD systems.


\bibliographystyle{apalike}
{\small \bibliography{paper.bib}}
\end{document}